\documentclass[11pt]{article}

\usepackage[final]{acl}

\usepackage{lmodern}
\usepackage{times}
\usepackage{latexsym}

\usepackage[T1]{fontenc}

\usepackage[utf8]{inputenc}

\usepackage{microtype}

\IfFileExists{inconsolata.sty}{\usepackage{inconsolata}}{}

\usepackage{graphicx}
\usepackage{multirow}
\usepackage{booktabs}
\usepackage{multicol}
\usepackage{amsmath}
\usepackage{fvextra}
\usepackage{float} 

\title{Decisions by Discrimination: Differential Evidence for Medical Reasoning}

\author{
 \textbf{Yunpeng Xiao\textsuperscript{1}},
 \textbf{Yuzhang Xie\textsuperscript{1}},
 \textbf{Junhao Huang\textsuperscript{1}},
 \textbf{Carl Yang\textsuperscript{1}},
 \textbf{Xiao Hu\textsuperscript{2}},
 \textbf{Kai Shu\textsuperscript{1}}
\\
 \textsuperscript{1}Department of Computer Science, Emory University,
\\
 \textsuperscript{2}Nell Hodgson Woodruff School of Nursing, Emory University
\\
{\fontfamily{qcr}\selectfont
  \shortstack[c]{
    \{yunpeng.xiao, yuzhang.xie, junhao.huang,\\
    j.carlyang, xiao.hu, kai.shu\}@emory.edu
  }
}
}

\hypersetup{pdftitle={Decisions by Discrimination: Differential Evidence for Medical Reasoning}}

\begin{document}
\maketitle
\begin{abstract}

Large language models (LLMs) are increasingly studied for clinical decision support, and medical question answering provides a testbed for evaluating this capability. In particular, medical multiple-choice questions (MCQs) reflect a core challenge in clinical decision-making: plausible options often share overlapping evidence, while only a few decisive factors distinguish the correct option from incorrect ones. However, existing retrieval-based and reasoning-oriented methods mainly focus on supplying broadly relevant knowledge or eliciting reasoning paths, rather than identifying the compact differential evidence that separates confusing alternatives. In this work, we present a retrieval-grounded differential evidence framework for medical MCQs. Our method extracts option-specific pieces of differential evidence and organizes them into concise evidence prompts for answer selection. This design better aligns external knowledge with differential reasoning while reducing irrelevant context. Experiments show consistent accuracy gains across multiple datasets and target LLMs. Ablation studies further show that our framework outperforms both directly using retrieved evidence chunks and closed-source LLM-generated general hints, and that its improvements are not solely explained by answer-proximal guidance, but are consistent with differential evidence serving as a compact aid for reasoning over competing alternatives. Our findings also offer insights for medical reasoning and other decision-making settings with multiple possible choices.

\end{abstract}

\section{Introduction}
Large language models (LLMs) are increasingly explored as clinical assistants, powering systems for triage, diagnosis support, and patient education \cite{singhal2023large, singhal2025toward, wu2024medjourney}. However, even medically tuned LLMs remain prone to unsafe reasoning behaviors. A common challenge in clinical reasoning is that many candidate diagnoses or treatment decisions share a large core of overlapping facts, while differing only in a small number of critical discriminatory factors \cite{bowen2006educational, cook2020higher}. If a model fails to identify and reason over these few distinguishing facts, then even access to abundant relevant knowledge may still lead to misdiagnosis or other incorrect clinical decisions.

In medical multiple-choice questions (medical MCQs), this challenge is especially well exposed, making them a natural benchmark for studying differential clinical reasoning. In real clinical practice, competing diagnoses or treatment options often share substantial overlapping evidence, with only a few decisive factors determining the correct choice. Medical MCQs capture this structure naturally: several answer options may each align with part of the vignette, but only one is supported by the discriminative evidence that resolves the case. This makes the benchmark useful for evaluating whether models can identify clinically meaningful differences among plausible hypotheses, rather than relying on broadly relevant but non-discriminative facts.

\begin{figure*}[htp]
    \centering
    \includegraphics[width=1\linewidth]{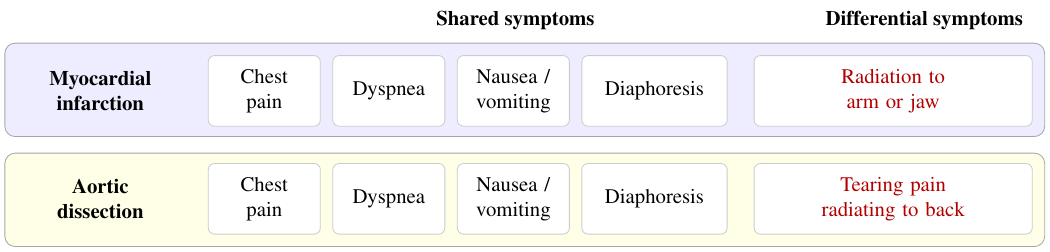}
    \caption{Two similar diseases may share most symptoms, while only a few discriminative symptoms determine the final diagnosis. Red text highlights the differential evidence.}
    \label{fig:example}
\end{figure*}

However, existing methods are still not well aligned with this requirement. Retrieval-augmented generation (RAG) has become a common strategy for medical QA, supplementing LLMs with external knowledge from structured resources such as knowledge graphs and unstructured corpora such as textbooks. Current medical RAG pipelines are primarily designed to retrieve generally relevant evidence, not the small set of discriminative facts that separates plausible answer options. As a result, they often return a broad supportive context while failing to surface the evidence needed to resolve near-miss alternatives. Figure \ref{fig:example} presents a concrete example in which two clinically similar diseases share multiple symptoms, while only a small number of distinctive symptoms determine the final diagnosis.

We therefore propose a retrieval-grounded differential evidence framework for medical MCQs. Instead of appending broad supporting context, our method extracts anchor cues, retrieves cue--option co-occurrence passages from textbooks, and distills them into compact option-specific evidence for answer selection. This design reduces retrieved context length by about 85\% while improving accuracy across benchmarks.

Our main contributions are:
\begin{itemize}
\item We formulate differential evidence as an evidence organization problem, emphasizing option-discriminative evidence rather than broadly relevant knowledge.

\item We propose a four-stage framework that extracts anchor cues, retrieves cue--option co-occurrence evidence, and distills it into compact option-specific evidence for answer selection.

\item Experiments on three medical MCQ benchmarks show consistent gains across multiple LLMs. Ablation studies with raw retrieved text, general hints, and evidence-only settings suggest that the gains come from differential evidence rather than more text or general answer cues.
\end{itemize}

\section{Related Work}
\subsection{Medical QA and Retrieval-Augmented Generation (RAG)}
Large-scale medical QA benchmarks, such as MedQA ~\cite{jin2021disease}, MedXpertQA ~\cite{zuo2025medxpertqa}, and MedBullets ~\cite{chen2025benchmarking}, have become important testbeds for evaluating fine-grained clinical reasoning in LLMs. In particular, multiple-choice medical QA provides a constrained decision-making setting in which several candidate options may all appear clinically plausible at first glance. This makes such benchmarks especially useful for assessing whether models can assist in real-world clinical decision making~\cite{chen2025benchmarking,wei2026epiqal,wang2025biomedjimpact}. 

Many medical QA tasks, especially MedQA-style problems, require reasoning beyond the information explicitly stated in the question stem, thus models often need to draw on background medical knowledge, including disease mechanisms, diagnostic criteria, treatment indications, and contraindications~\cite{lievin2024can,lucas2024reasoning,bhasuran2025preliminary}. Since this knowledge is broad, evolving, and difficult to capture reliably through parametric memory alone, retrieval-augmented generation (RAG) over external medical knowledge bases has emerged as a natural solution~\cite{yang2025retrieval}. By retrieving evidence from curated resources such as clinical guidelines, biomedical ontologies, textbooks, and knowledge graphs, medical RAG systems can ground answer selection in explicit evidence rather than latent memorization~\cite{amugongo2025retrieval,ke2025retrieval,song2025graph}. Recent work has explored this direction through methods such as MIRAGE~\cite{xiong2024mirage}, K-Comp~\cite{cho2025k-comp}, MKRAG~\cite{shi2025mkrag}, MedRAG~\cite{zhao2025medrag}, BriefContext~\cite{zhang2025briefcontext}, RAG$^2$\cite{sohn2025rationale}, KGARevion~\cite{su2024kgarevion}, and Self-BioRAG~\cite{jeong2024selfbiorag}. In addition to traditional text-based RAG, recent studies have explored KG-guided retrieval and reasoning frameworks, such as Dr.Knows~\cite{gao2025LLMKG}, MedIKAL~\cite{jia2025LLMKG}, KERAP~\cite{xie2025kerap}, KG4Diagnosis~\cite{zuo2024LLMKG}, and KGARevion~\cite{su2024kgarevion}, which leverage knowledge graphs to facilitate answer generation, diagnostic reasoning, and evidence retrieval.

Although retrieved external evidence can improve factual grounding, conventional retrieval generally prioritizes overall semantic relevance to the question, often returning broad background knowledge that may support multiple candidate options simultaneously~\cite{amugongo2025retrieval,liu2025improving,xiong2024improving}. However, for medical multiple-choice reasoning, the key challenge is often not general relevance but discriminative utility: the retrieved evidence should help the model determine why one option is more appropriate than other clinically plausible alternatives under the specific case context~\cite{chen2023distinguish,singh2025options,alhazmi2024distractor}. These observations make differential evidence especially well-suited for medical QA. 

\subsection{Option Discrimination and Evidence Selection in Multiple-Choice Reasoning}
Existing work has increasingly suggested that the main challenge in multiple-choice reasoning is not merely identifying generally relevant knowledge, but distinguishing among candidate options that may all appear partially plausible. Early work on multi-choice reading comprehension emphasized joint interactions among the passage, question, and candidate answers, rather than question-only matching \cite{wang2018co}. Recent studies further show that LLMs are sensitive to candidate-option structure: some methods model elimination over options \cite{ma2023poe}, while others find that predictions can vary under answer-order permutations, indicating unstable discrimination among alternatives \cite{pezeshkpour2024large}. Related analyses also show that benchmarks often contain several plausible candidates, and that models may select the least incorrect option rather than one that is distinctly correct \cite{wang2025llms, palta2024plausibly}. This issue is especially salient in medical MCQs, where options often share overlapping clinical facts and differ only in a few decisive mechanisms, contraindications, or management roles. In parallel, retrieval-based methods have incorporated answer options into retrieval or knowledge selection, such as question-aware knowledge graph prompting and answer-conditioned support retrieval \cite{zhang2025retrieving, liu2025question}. However, these methods mainly improve option-aware knowledge access, while compact differential evidence for option-level comparison remains underexplored. Unlike prior option-aware retrieval methods that primarily improve access to option-related knowledge, our focus is on organizing knowledge into differential evidence that clearly distinguishing between the various options.

\section{Methods}
\begin{figure*}[htp]
    \centering
    \includegraphics[width=1.\linewidth]{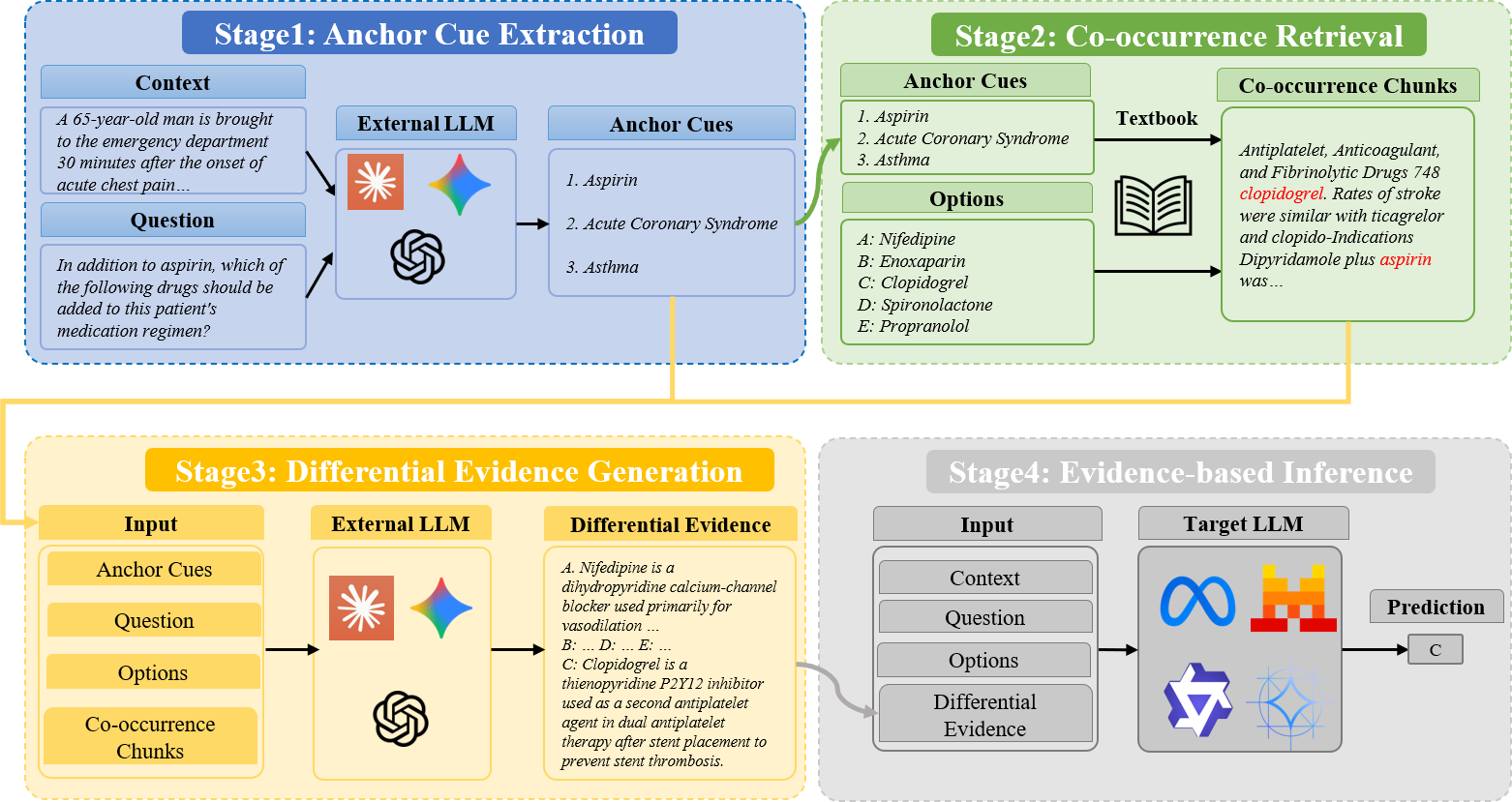}
    \caption{The framework of our proposed method.}
    \label{fig:Mthod}
\end{figure*}

In this section, we will introduce the proposed differential evidence reasoning method for clinical decision-making, especially in multiple-choice questions. Figure \ref{fig:Mthod} shows the framework of our proposed method, which consists of four stages: anchor cue extraction, cue-option co-occurrence retrieval, differential evidence generation, and evidence-based inference. The first three stages form the differential evidence construction pipeline, while the fourth stage uses the generated evidence for final answer prediction.

\subsection{Definition of Differential Evidence}

Given a multiple-choice question with candidate options
\[
O = \{o_1, o_2, \dots, o_m\},
\]
we define differential evidence for an option \(o_i\) as external evidence that supports \(o_i\) more strongly than competing options under the current case context. In other words, differential evidence is not merely relevant knowledge, but knowledge with discriminative utility for separating clinically plausible alternatives.

Formally, if \(s(e,o)\) denotes the degree to which an evidence item \(e\) supports option \(o\), then ideal differential evidence for \(o_i\) would satisfy
\[
s(e,o_i) > s(e,o_k), \quad \forall k \neq i,
\]
or more generally provide substantially stronger support for \(o_i\) than for the strongest competing option.

In practice, explicit support scoring is difficult because raw textbook passages are often long, redundant, and only indirectly aligned with option-level distinctions. We therefore approximate differential evidence through a retrieval-and-distillation pipeline. The following subsections describe the first three stages of the framework for differential evidence construction, while the fourth stage performs evidence-based inference.

\subsection{Definition of Atomic Facts and Anchor Cues}

In medical multiple-choice questions, one data sample typically consists of the parts: the background context $C$, the explicit question $Q$ and the options $O$. Since the key clinical evidence is usually conveyed in the background description, we decompose $C$ into a set of atomic facts, following the intuition of fine-grained factual decomposition in fact-checking \cite{min2023factscore}:
\[C = \{f_1, f_2, \ldots, f_n\},\]
where each $f_i$ denotes a minimal clinical fact, such as a symptom, laboratory finding, medical history element, or treatment detail.

However, atomic decomposition alone is insufficient, as the fact set may still contain substantial redundancy and many facts that are only weakly relevant to the question $Q$ and final decision. We therefore define anchor cue, which is to distill the original fact set into a compact set of highly salient cues. Formally, we select a subset of atomic facts,
\[A = \{a_1, a_2, \ldots, a_i\} \subseteq C,\quad i \ll n,\]

where $A$ denotes the set of anchor cues.

This step can be understood as a general content distillation problem in text understanding: from a larger collection of fine-grained facts, we aim to retain only those that best preserve the core semantics and decision-critical information of the scenario. This step is not intended for full clinical reasoning or direct answer prediction, but to distill the scenario into a few salient anchor cues, similar to focused reading comprehension over clinical text. For example, the context mentions that \emph{the patient developed tinnitus} while the question asks \emph{Which medication caused this phenomenon}, then "tinnitus" serves as the anchor cue for this specific question. 

\subsection{Textbook Knowledge Retrieval through Anchor Cue–Option Co-occurrence}
Similar to traditional RAG, we also need to leverage external knowledge to enhance the performance of the target model in medical scenarios. However, as noted in Section 2, current knowledge graphs are not yet sufficient to distinguish the subtle nuances between options and generate discriminative evidence; therefore, we identify chunks in textbooks that exhibit co-occurrence between anchor cues and the options as our preliminary source of external knowledge.

Specifically, by leveraging the powerful semantic understanding capabilities of external LLMs, we first input the context $C$ and the question $Q$ into these models to extract anchor cues \[A = \{a_1, a_2, \ldots, a_i\},\] and normalize these cues and generate their common variants (e.g., singular/plural forms, abbreviations, etc.). Similarly, we input the options $O$ into the external LLMs, subjecting them to the same normalization process and generating their common variants \[O_n = \{o_1, o_2, \ldots, o_j\}.\] This strict input separation reduces the risk that external LLMs act as implicit solvers of the task. Instead, their role is limited to localized semantic normalization and variant generation, rather than answer inference over the full MCQ instance.

Next, we use the textbook provided in the MedQA dataset \cite{jin2021disease} to identify co-occurring chunks \[p_{ij} = \text{Co-occurr}(a_i,o_j).\] 
A chunk is considered co-occurring if it contains at least one normalized variant of an anchor cue and one normalized variant of an option concept. In this work, to control prompt length and reduce redundant evidence, we extract only one chunk for each anchor cue-option pair. Specifically, we extract a maximum of $|A| \times |O|$ paragraphs. More details are shown in appendix.

\subsection{Textbook-Constrained Differential Evidence Generation}
Compared with knowledge graphs, textbook knowledge provides substantially richer semantic content. However, it is often fragmented, redundant, and not explicitly organized around option-level distinctions, which makes it difficult to directly obtain discriminative evidence for specific answer choices. In addition, textbook retrieval may still miss relevant information due to limited coverage or retrieval failures. To address this, we use external LLMs to generate option-discriminative evidence while treating retrieved textbook knowledge as a constraint on the generation process. Formally, we generate the set of differential evidence
$D = \{d_A, d_B, ...\}$ where $d_A$ denotes the evidence specific to Option A, as follows:

\[
    D = ExternalLLM(A, Q, O, P).\]

Here, $A$ is the set of anchor cues, $O$ is the set of original options, $P = \bigcup {p_{ij}}$ is the set of retrieved external knowledge, and $Q$ is the question. Each $d_o$ is generated as a short, option-specific, condition-oriented statement capturing the conditions that most clearly distinguish option from competing alternatives. When retrieved evidence is missing for a given option, the LLM is allowed to provide limited supplementation from internal knowledge.

\subsection{Differential Evidence Quality Control}

Because external LLMs may have been pretrained on overlapping medical QA data, we introduce several safeguards to reduce answer leakage and bias. First, during anchor cue extraction and normalization, the external LLM only sees the context and question $C,Q$ without the options $O$; during option extraction and normalization, it only sees the options $O$ without the context and question $C,Q$; during differential evidence generation, it only sees the anchor cues, questions, options $A,Q,O$ without context $C$. As a result, the external model never has access to the full MCQ instance at any stage. Second, the system prompt explicitly instructs the model to prioritize retrieved textbook evidence and only supplement missing details when retrieval is insufficient. Finally, for differential evidence generation with external LLMs, we use carefully designed prompts to encourage the production of option-discriminative evidence while preserving as much neutrality as possible.

\subsection{Differential Evidence-Based Inference}
Upon obtaining the differential evidence set, the target LLMs—typically distinct from the external LLM used in the preceding step—infer the answers to the medical MCQs using the following method:

\[Answer = TargetLLM(C, Q, O, D).\]

Here, $C$ is the clinical context, $Q$ is the question, $O$ is the set of original options, and $D$ is the generated differential evidence. The target LLM takes these components as input and produces the final answer.

\section{Experiments}
In this section, we will introduce the dataset used in the experiments, the LLMs employed to generate anchor cues and evidence (External LLMs), and the LLMs used for testing (Target LLMs). We then answer the following research questions:

\begin{itemize}
\item RQ1: Can differential evidence improve model performance?
\item RQ2: Does differential evidence provide benefits beyond generic auxiliary evidence?
\item RQ3: Does differential evidence genuinely support reasoning, or does it risk improving performance by providing answer-proximal guidance?
\end{itemize}

\subsection{Datasets and Models}

\begin{table*}[!htp]
\centering
\small
\setlength{\tabcolsep}{2pt}
\begin{tabular}{lcccccccccccc}
\toprule
\multirow{2}{*}{\textbf{Model}} 
& \multicolumn{4}{c}{\textbf{MedQA (5-op)}} 
& \multicolumn{4}{c}{\textbf{MedBullets (5-op)}} 
& \multicolumn{4}{c}{\textbf{MedXpert-Text}} \\
\cmidrule(lr){2-5} \cmidrule(lr){6-9} \cmidrule(lr){10-13}
& \textbf{Base} & \textbf{CoT} & \textbf{Diff.} & \shortstack{\textbf{CoT}\\\textbf{+Diff.}}
& \textbf{Base} & \textbf{CoT} & \textbf{Diff.} & \shortstack{\textbf{CoT}\\\textbf{+Diff.}}
& \textbf{Base} & \textbf{CoT} & \textbf{Diff.} & \shortstack{\textbf{CoT}\\\textbf{+Diff.}} \\
\midrule
Llama-3-8B     & 52.40 & 65.02 & 65.12 & \textbf{67.01} & 44.16 & 49.35 & 50.00 & \textbf{60.06} & 11.31 & 14.73 & 14.53 & \textbf{18.20} \\
Llama-3-70B    & 74.31 & 79.26 & 79.81 & \textbf{81.70} & 61.36 & 67.86 & 68.83 & \textbf{68.83} & 21.63 & 23.68 & 23.84 & \textbf{23.88} \\
Gemma-3-27B-It & 64.02 & 71.25 & 71.33 & \textbf{75.81} & 46.75 & 50.00 & 51.95 & \textbf{53.57} & 14.61 & 15.63 & 16.98 & \textbf{19.27} \\
Qwen-3-32B     & 69.91 & 73.68 & 75.33 & \textbf{78.95} & 55.19 & 60.06 & 58.44 & \textbf{64.61} & 17.14 & 18.82 & 19.55 & \textbf{21.47} \\
Mistral-3-8B   & 56.24 & 68.97 & 64.96 & \textbf{74.16} & 42.86 & 50.97 & 50.97 & \textbf{59.09} & 15.39 & 16.86 & 17.76 & \textbf{19.35} \\
\midrule
\textbf{Average} & 63.38 & 71.64 & 71.31 & \textbf{75.53} & 50.06 & 55.65 & 56.04 & \textbf{61.23} & 16.02 & 17.94 & 18.53 & \textbf{20.43} \\
\bottomrule
\end{tabular}
\caption{Performance comparison across models, datasets, and prompting settings. Base denotes direct answering without differential evidence, CoT denotes zero-shot chain-of-thought prompting, Diff.\ denotes augmentation with differential evidence, and CoT+Diff.\ combines chain-of-thought prompting with differential evidence. All values are reported in accuracy (\%).}
\label{tab:main_results}
\end{table*}

\textbf{Datasets.} We evaluate our method on MedQA \cite{jin2021disease}, MedXpertQA-Text \cite{zuo2025medxpertqa}, and MedBullets \cite{chen2025benchmarking} because together they provide a strong testbed for scenario-grounded, option-discriminative medical reasoning. By contrast, some medical QA benchmarks, such as PubMedQA \cite{jin2019pubmedqa} and MedMCQA \cite{pal2022medmcqa}, generally feature shorter question stems and are more oriented toward knowledge-centric medical question answering. Therefore, they are less suitable for evaluating the scenario-grounded differential reasoning targeted in this work.

\textbf{External LLMs}. Motivated by RAG$^2$ \cite{sohn2025rationale}, we view the external LLM as a differential evidence generator. Its quality mainly affects the generated evidence, which then influences downstream reasoning. We tested several closed-source LLMs, including Claude-family models, and selected GPT-5 mini \cite{singh2025openai} as the evidence generator because it offered the best overall trade-off among evidence quality, instruction following, cost, and API stability. To improve reproducibility and assess dependence on closed-source models, we also tested open-source models. Among them, DeepSeek-V3.2 \cite{liu2025deepseek} produced the most reliable differential evidence in our experiments. We therefore report an additional ablation using DeepSeek-V3.2 as the evidence generator.

\textbf{Target LLMs}. We evaluate our method on a series of open-source LLMs. These include Llama 3 \cite{grattafiori2024llama}, Mistral \cite{mistral2025mistral3}, Gemma \cite{team2025gemma3} and Qwen 3 \cite{yang2025qwen3}. All experiments are conducted through Amazon Bedrock under a unified inference setup, using deterministic decoding. More details are shown in appendix.

\subsection{Performance of Differential Evidence (RQ1)}
We use four settings for the experiments: \textbf{Baseline}, \textbf{CoT}, \textbf{Differential Evidence} and \textbf{CoT + Differential Evidence} . 
\textbf{Baseline} directly predicts the answer from the original context, question, and options. 
\textbf{CoT} uses the same input with an additional zero-shot chain-of-thought instruction. 
\textbf{Differential Evidence} further augments the input with our generated option-specific differential evidence.
\textbf{CoT + Differential Evidence} setting simultaneously utilize CoT and differential evidence.

As shown in Table ~\ref{tab:main_results}, the \textbf{CoT + Differential Evidence} setting achieves better performance than both \textbf{Baseline} and \textbf{CoT} across all five models. Notably, using \textbf{Differential Evidence} alone already achieves performance close to \textbf{CoT} in many settings, indicating that the generated evidence itself provides useful external knowledge and option-discriminative information.

We further examine whether the improvement is tied to a single proprietary generator by replacing GPT-5 mini with DeepSeek-V3.2 as the evidence generator. Table~\ref{tab:generator_ablation} shows that DeepSeek-generated evidence improves all five target models, increasing the average MedQA accuracy from 63.38\% to 69.09\%. While GPT-5 mini remains stronger on average, the consistent gains from DeepSeek-V3.2 indicate that the benefits of differential evidence are not exclusive to one generator, though generator quality still affects the magnitude of improvement.

\begin{table}[t]
\centering
\small
\setlength{\tabcolsep}{3pt}
\begin{tabular}{lccc}
\toprule
Model & Base & \shortstack{+Diff.\\DeepSeek} & \shortstack{+Diff.\\GPT-5 mini} \\
\midrule
Llama-3-8B       & 52.40 & 61.19 & 65.12 \\
Llama-3-70B      & 74.31 & 79.96 & 79.81 \\
Gemma-3-27B-It   & 64.02 & 68.97 & 71.33 \\
Qwen-3-32B       & 69.91 & 73.29 & 75.33 \\
Mistral-3-8B     & 56.24 & 62.06 & 64.96 \\
\midrule
Average          & 63.38 & 69.09 & 71.31 \\
\bottomrule
\end{tabular}
\caption{Evidence generator ablation on MedQA (5-option). We compare differential evidence generated by GPT-5 mini and DeepSeek-V3.2 under the non-CoT setting.}
\label{tab:generator_ablation}
\end{table}

\subsection{Benefits Beyond Generic Retrieved Evidence and Generic Hints (RQ2)}
To examine whether the gains of differential evidence go beyond simply adding more external text, we compare two CoT-based settings: \textbf{CoT + Text}, which directly appends the retrieved co-occurrence paragraphs to the input, and \textbf{CoT + Differential Evidence}, which instead provides the model with compact option-specific differential evidence distilled from those retrieved materials.

As shown in Table ~\ref{tab:rq2_rag_vs_diff}, directly using the retrieved co-occurrence text does not reliably improve performance. In many cases, \textbf{CoT + Text} even performs worse than the CoT baseline, especially on MedQA and MedBullets. By contrast, \textbf{CoT + Differential Evidence} consistently yields positive gains over CoT across all three datasets. This contrast suggests that the benefit does not come from merely exposing the model to more retrieved content, but from organizing external knowledge into a concise and discriminative form that is more aligned with multiple-choice decision making.

\begin{table*}[t]
\centering
\small
\setlength{\tabcolsep}{2pt}
\begin{tabular}{lcccccccccccc}
\toprule
\multirow{2}{*}{\textbf{Model}} 
& \multicolumn{4}{c}{\textbf{MedQA (5-op)}} 
& \multicolumn{4}{c}{\textbf{MedBullets (5-op)}} 
& \multicolumn{4}{c}{\textbf{MedXpert-Text}} \\
\cmidrule(lr){2-5} \cmidrule(lr){6-9} \cmidrule(lr){10-13}
& \shortstack{\textbf{CoT}\\\textbf{+Text}} & \textbf{$\Delta$} & \shortstack{\textbf{CoT}\\\textbf{+Diff.}} & \textbf{$\Delta$}
& \shortstack{\textbf{CoT}\\\textbf{+Text}} & \textbf{$\Delta$} & \shortstack{\textbf{CoT}\\\textbf{+Diff.}} & \textbf{$\Delta$}
& \shortstack{\textbf{CoT}\\\textbf{+Text}} & \textbf{$\Delta$} & \shortstack{\textbf{CoT}\\\textbf{+Diff.}} & \textbf{$\Delta$} \\
\midrule
Llama-3-8B     & 58.84 & -6.18  & \textbf{67.01} & \textbf{+1.99} & 42.53 & -6.82  & \textbf{60.06} & \textbf{+10.71} & 14.78 & +0.05  & \textbf{18.20} & \textbf{+3.47} \\
Llama-3-70B    & 75.73 & -3.53  & \textbf{81.70} & \textbf{+2.44} & 56.17 & -11.69 & \textbf{68.83} & \textbf{+0.97}  & 22.69 & -0.99  & \textbf{23.88} & \textbf{+0.20} \\
Gemma-3-27B-It & 65.51 & -5.74  & \textbf{75.81} & \textbf{+4.56} & 49.03 & -0.97  & \textbf{53.57} & \textbf{+3.57}  & 15.14 & -0.49  & \textbf{19.27} & \textbf{+3.64} \\
Qwen-3-32B     & 73.29 & -0.39  & \textbf{78.95} & \textbf{+5.27} & 53.25 & -6.81  & \textbf{64.61} & \textbf{+4.55}  & 19.92 & +1.10  & \textbf{21.47} & \textbf{+2.65} \\
Mistral-3-8B   & 58.37 & -10.60 & \textbf{74.16} & \textbf{+5.19} & 45.45 & -5.52  & \textbf{59.09} & \textbf{+8.12}  & 18.04 & +1.18  & \textbf{19.35} & \textbf{+2.49} \\
\midrule
\textbf{Average} & 66.35 & -5.29 & \textbf{75.53} & \textbf{+3.89} & 49.29 & -6.36 & \textbf{61.23} & \textbf{+5.58} & 18.11 & +0.17 & \textbf{20.43} & \textbf{+2.49} \\
\bottomrule
\end{tabular}
\caption{Comparison between generic retrieved text and differential evidence under the same chain-of-thought (CoT) reasoning framework. $\Delta$ over CoT denotes the absolute accuracy gain relative to the CoT baseline in Table~\ref{tab:main_results}.}
\label{tab:rq2_rag_vs_diff}
\end{table*}

We further compare the output length of the two methods using word count. As shown in Table~\ref{tab:length_comparison}, differential evidence is substantially more compact than directly concatenated retrieved text across all three datasets. For example, on MedQA the average length decreases from 677.3 words for co-occurrence text to 108.5 words for differential evidence, with similar reductions on MedBullets and MedXpertQA-Text. This suggests that the gain of our framework comes not from adding more retrieved content, but from organizing external knowledge into a shorter and more discriminative form. Moreover, the substantial number of non-empty co-occurrence retrievals suggests that our method is supported by sufficient retrieved paragraphs, rather than depending purely on the external LLM to fabricate evidence.

\begin{table*}[t]
\centering
\small
\setlength{\tabcolsep}{5pt}
\begin{tabular}{lccccc}
\toprule
Dataset & Diff. Mean & Diff. Median & Co-occ. Mean & Co-occ. Median & Non-empty \\
\midrule
MedQA      & 108.5 & 108.0 & 677.3  & 571.0 & 1023/1273 \\
MedBullets & 112.2 & 112.5 & 734.6  & 621.5 & 264/308 \\
MedXpert   & 207.1 & 207.0 & 1160.5 & 957.0 & 1829/2450 \\
\bottomrule
\end{tabular}
\caption{Length comparison in words between co-occurrence text and differential evidence. Non-empty counts questions with at least one retrieved co-occurrence passage.}
\label{tab:length_comparison}
\end{table*}

These findings indicate that generic retrieved text can be too long, noisy, or weakly structured to effectively support reasoning, and may even distract the model from the decisive clues in the question. This observation is also consistent with prior studies reporting that directly concatenating retrieved medical text often provides limited benefit and can sometimes degrade performance due to irrelevant or diffuse context \cite{kim2025rethinking}. In comparison, differential evidence appears to make better use of external knowledge by highlighting the most distinguishing facts for separating competing options, rather than requiring the model to infer these distinctions from raw retrieved paragraphs on its own.

To test whether the gains come merely from adding LLM-generated auxiliary medical text, we construct a general-hint baseline using the same GPT-5 mini generator. This baseline is conditioned on the same anchor cues, question, and answer options, but it does not use retrieved textbook co-occurrence chunks and is not instructed to identify option-discriminative conditions. Instead, it asks the generator to produce general option-level medical background notes. 

As shown in Table~\ref{tab:general_hint}, general hints provide only a modest average improvement over the base setting, increasing MedQA accuracy from 63.38\% to 64.71\%. In contrast, differential evidence improves the average accuracy to 71.31\%. This gap suggests that the benefit of our method is not simply due to appending additional LLM-generated text, but comes from organized differential evidence.

\begin{table}[t]
\centering
\small
\setlength{\tabcolsep}{3pt}
\begin{tabular}{lccc}
\toprule
Model & Base & \shortstack{+General\\Hint} & \shortstack{+Diff.\\Evidence} \\
\midrule
Llama-3-8B       & 52.40 & 53.57 & 65.12 \\
Llama-3-70B      & 74.31 & 77.22 & 79.81 \\
Gemma-3-27B-It   & 64.02 & 65.28 & 71.33 \\
Qwen-3-32B       & 69.91 & 69.99 & 75.33 \\
Mistral-3-8B     & 56.24 & 57.50 & 64.96 \\
\midrule
Average          & 63.38 & 64.71 & 71.31 \\
\bottomrule
\end{tabular}
\caption{Comparison with GPT-5-mini-generated general hints on MedQA (5-option).}
\label{tab:general_hint}
\end{table}

\subsection{Reasoning Support or Answer-Proximal Guidance? (RQ3)}

\begin{figure*}[t]
    \centering
    \includegraphics[width=1\linewidth]{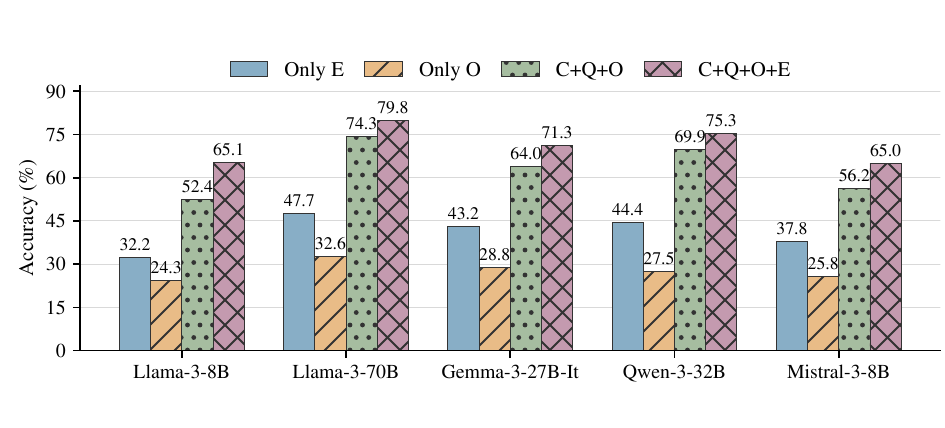}
    \caption{Comparison of four input settings on MedQA for RQ3: Only options (Only O), Only evidence (Only E), context+question+options (C+Q+O), and context+question+options+evidence (C+Q+O+E). Differential evidence alone provides useful signal, but yields the strongest gains when combined with the original question context and candidate options.}
    \label{fig:rq3_medqa}
\end{figure*}

We further investigate whether differential evidence improves performance by genuinely supporting reasoning, or by introducing answer-proximal guidance (e.g., the evidence includes a sentence like ``The answer is ...''). To this end, we evaluate four input settings on three datasets: only options (Only $O$), only evidence (Only $E$), context + question + options ($C{+}Q{+}O$), and context + question + options + evidence ($C{+}Q{+}O{+}E$). 

Among these settings, $C{+}Q{+}O$ serves as the standard baseline, while $C{+}Q{+}O{+}E$ tests whether differential evidence provides additional value beyond the original question input. The Only $O$ and Only $E$ settings are included to analyze how much predictive signal is contained in the options and the generated evidence, respectively. Specifically, Only $O$ measures how much answer signal is already exposed by the candidate options themselves, while Only $E$ tests whether the generated evidence alone is overly predictive of the correct answer. For clarity, we visualize the MedQA results in the main paper and report the full results on all three datasets in Appendix.

As shown in Figure ~\ref{fig:rq3_medqa}, the evidence-only setting achieves clearly above-chance performance, indicating that differential evidence carries useful answer-relevant information. However, its performance remains substantially below the full-input setting. On average, Only $E$ reaches 41.05\% accuracy, whereas $C{+}Q{+}O{+}E$ reaches 71.31\%, a gap of more than 30 percentage points. This suggests that differential evidence alone is not sufficient to solve the questions reliably in isolation.

These results suggest that the gains from differential evidence are not solely explained by direct answer cues. While the evidence itself is informative, it is not sufficient to fully solve the questions in isolation. 

We emphasize that differential evidence is expected to be answer-relevant; otherwise it would not be useful evidence. The goal of this analysis is therefore not to show that evidence contains no predictive signal, but to test whether it collapses into an answer shortcut.

\section{Discussion and Conclusion}
We present a differential evidence framework for medical multiple-choice question answering. Experiments across three medical QA benchmarks and multiple target models show that differential evidence consistently improves direct answering and further complements chain-of-thought prompting. Comparisons with general hints and retrieved chunks suggest that the gains are not merely due to adding extra generated text, but depend on the quality and organization of the evidence. Leakage-oriented analyses further indicate that differential evidence is informative but does not function as a standalone answer channel. Although our experiments focus on medical MCQs, the underlying idea may apply more broadly to decision-making settings where candidate hypotheses, plans, or actions are explicitly provided or can be enumerated, and where a small number of discriminative facts determine the final choice. 

\section*{Limitations}
Our study still has some limitations. First, our framework depends on the quality of the external evidence generator. Although GPT-5 mini and DeepSeek-V3.2 produce useful differential evidence in our experiments, weaker generators may produce overly generic, overly short, or answer-proximal statements, reducing downstream gains. In addition, while our prompts and leakage analyses aim to prevent the evidence from directly revealing the answer, subtle answer-proximal cues may still occur, particularly in management and contraindication questions. Finally, our experiments are limited to medical multiple-choice benchmarks with explicit candidate options, and extending the framework to open-ended or real-world decision settings will require additional validation.

\section*{Ethical Considerations}

This work evaluates evidence-based answering on medical multiple-choice benchmarks. Benchmark accuracy does not establish clinical safety or effectiveness. Generated evidence, including supplementation from a model's internal knowledge, may be incomplete, inaccurate, or biased. These outputs should not be used as patient-specific medical advice or as a substitute for qualified clinical judgment; clinical use would require independent expert review and validation.

We used GPT-5 for limited drafting and language editing assistance during manuscript preparation. All technical content, experimental design, analysis, and final wording were reviewed and verified by the authors, who take full responsibility for the final paper.

\section*{Acknowledgements}

This material is based upon work supported by NSF awards (IIS-2506643, and POSE-2346158). It was partially supported by the internal funds and GPU servers provided by the Computer Science Department of Emory University, the US NSF Award Numbers 2442172, 2312502, 2319449, and the US National Institutes of Health under Award Numbers K25DK135913, RF1NS139325, R01DK143456, U18DP006922 and R01EB038426. The views and conclusions contained in this document are those of the authors and should not be interpreted as necessarily representing the official policies, either expressed or implied, of the National Science Foundation.

\bibliography{custom}

\appendix

\section{Additional Detailed Results for RQ3}

We report the full RQ3 results on all three datasets in Appendix Table~\ref{tab:rq3_full_results}. We observe that on MedXpert-Text, the evidence-only setting is relatively strong and for some models is comparable to, or slightly better than, $C{+}Q{+}O$. A possible reason is that the overall baseline performance on this dataset is low, suggesting that the task is difficult for the tested models under the standard input. In such cases, even a small amount of compact external signal may lead to a noticeable gain. However, $C{+}Q{+}O{+}E$ still yields the strongest overall results, indicating that differential evidence is most effective when combined with the original context.

\begin{table*}
\centering
\small
\setlength{\tabcolsep}{2pt}
\begin{tabular}{lcccccccccccc}
\toprule
\multirow{2}{*}{\textbf{Model}}
& \multicolumn{4}{c}{\textbf{MedQA (5-op)}}
& \multicolumn{4}{c}{\textbf{MedBullets (5-op)}}
& \multicolumn{4}{c}{\textbf{MedXpert-Text}} \\
\cmidrule(lr){2-5} \cmidrule(lr){6-9} \cmidrule(lr){10-13}
& \shortstack{\textbf{Only}\\\textbf{$O$}} & \shortstack{\textbf{Only}\\\textbf{$E$}} & \textbf{Base} & \textbf{+Diff.}
& \shortstack{\textbf{Only}\\\textbf{$O$}} & \shortstack{\textbf{Only}\\\textbf{$E$}} & \textbf{Base} & \textbf{+Diff.}
& \shortstack{\textbf{Only}\\\textbf{$O$}} & \shortstack{\textbf{Only}\\\textbf{$E$}} & \textbf{Base} & \textbf{+Diff.} \\
\midrule
Llama-3-8B     & 24.27 & 32.21 & 52.40 & \textbf{65.12} & 23.05 & 35.71 & 44.16 & \textbf{50.00} & 13.14 & 14.61 & 11.31 & \textbf{14.53} \\
Llama-3-70B    & 32.60 & 47.68 & 74.31 & \textbf{79.81} & 30.52 & 39.61 & 61.36 & \textbf{68.83} & 16.04 & 17.92 & 21.63 & \textbf{23.84} \\
Gemma-3-27B-It & 28.83 & 43.21 & 64.02 & \textbf{71.33} & 29.55 & 37.66 & 46.75 & \textbf{51.95} & 14.08 & 18.37 & 14.61 & \textbf{16.98} \\
Qwen-3-32B     & 27.49 & 44.38 & 69.91 & \textbf{75.33} & 27.60 & 41.23 & 55.19 & \textbf{58.44} & 13.22 & 15.06 & 17.14 & \textbf{19.55} \\
Mistral-3-8B   & 25.84 & 37.78 & 56.24 & \textbf{64.96} & 26.62 & 36.04 & 42.86 & \textbf{50.97} & 12.73 & 16.08 & 15.39 & \textbf{17.76} \\
\midrule
\textbf{Average} & 27.81 & 41.05 & 63.38 & \textbf{71.31} & 27.47 & 38.05 & 50.06 & \textbf{56.04} & 13.84 & 16.41 & 16.02 & \textbf{18.53} \\
\bottomrule
\end{tabular}
\caption{Full RQ3 results on all three datasets. Base denotes $C{+}Q{+}O$, and +Diff. denotes $C{+}Q{+}O{+}E$. We report accuracy (\%) under four input settings: only options (Only $O$), only evidence (Only $E$), context+question+options ($C{+}Q{+}O$), and context+question+options+evidence ($C{+}Q{+}O{+}E$). Here, $C{+}Q{+}O$ and $C{+}Q{+}O{+}E$ correspond to the \textbf{Base} and \textbf{Diff.} settings in the main results table, respectively.}
\label{tab:rq3_full_results}
\end{table*}

\section{Co-occurrence Retrieval Details}
For external knowledge retrieval, we use the textbook corpus provided together with the MedQA dataset \cite{jin2021disease}. Besides the question data, MedQA also releases reference textbooks that have already been segmented into chunks, and we directly perform co-occurrence retrieval over these pre-defined chunks. Since some chunks are overly long, we additionally cap the length of each retrieved chunk at 400 words. When multiple chunks satisfy the co-occurrence criterion for the same anchor cue-option pair, we keep only the first retrieved chunk.

We also explored several retrieval variants, including BM25-based retrieval strategies for identifying anchor-option co-occurrence paragraphs. In pilot experiments, these variants produced similar downstream trends. Since retrieval in our framework mainly serves as a grounding constraint for evidence distillation rather than as the final answer-selection mechanism, we use a simple and reproducible retrieval strategy in the main experiments.

\section{Implementation Details}

We use a unified inference setup for all target models. For direct answer generation, we set \emph{max new tokens} =16; for CoT-based settings, we set \emph{max new tokens} =512 to allow sufficient space for intermediate reasoning. For decoding, we use \emph{temperature} =0.0 and \emph{top p}=1.0 across all experiments. 

\section{Prompts}

\subsection{Anchor Cue Extraction and Normalization}
\begin{Verbatim}[fontsize=\small,breaklines=true,breakanywhere=true,breaksymbolleft={},breaksymbolright={},frame=single,framesep=3pt]
SYSTEM
You are an extractor for USMLE/MedQA-style question stems.

USER
You are an extractor for USMLE/MedQA-style question stems.

Input:
- "Question": the full question stem

Task:
Identify:
1) `question`: the actual question being asked, usually the final sentence
2) `entities`: up to 3 medical entities extracted only from the context/background part

Rules:
1) Use only the input stem.
2) Do not infer from answer choices.
3) Do not hallucinate image findings.
4) Do not output the context itself.
5) Extract entities only from the context, not from the final question sentence.
6) Return at most 3 entities, choosing the most answer-relevant ones.
7) Prefer concise, concrete, medically meaningful entities.
8) Each entity must contain `text`, `normalized`, and `variants`.
9) `variants` should contain up to 2 common alternate forms only, generally includes singular, plural, and abbreviation.
10) Deduplicate entities.

Example:
Input Question:
"A 25-year-old woman has fever, dysuria, and urinary frequency for 2 days. Which of the following is the most likely diagnosis?"

Output:
{
  "question": "Which of the following is the most likely diagnosis?",
  "entities": [
    {"text": "fever", "normalized": "fever", "variants": []},
    {"text": "dysuria", "normalized": "dysuria", "variants": []},
    {"text": "urinary frequency", "normalized": "urinary frequency", "variants": []}
  ]
}

Return structured JSON matching the schema exactly.

[paste context and question]
\end{Verbatim}

\subsection{Option Normalization}
\begin{Verbatim}[fontsize=\small,breaklines=true,breakanywhere=true,breaksymbolleft={},breaksymbolright={},frame=single,framesep=3pt]
SYSTEM
You are an extractor for USMLE/MedQA-style question stems.

USER
You will receive a JSON object with:
- "options": a dictionary of answer options, such as {"A": "...", "B": "...", ...}

Your task is to normalize the medical entities appearing in EACH option separately.

GOAL:
For each option, extract the core medical entity or entities explicitly present in that option text,
and provide:
- text
- normalized
- variants

IMPORTANT RULES:
1) Use ONLY the option text provided.
2) Do NOT use the question stem.
3) Do NOT infer hidden meaning beyond what is explicitly written in the option.
4) For a simple option such as a drug, disease, or procedure name, usually extract just one entity.
5) Prefer concise, text-grounded entity spans.
6) Deduplicate entities within each option.
7) `normalized` should be a concise canonical medical form (lower-case preferred).
8) `variants` should contain up to 2 common alternate names/forms only, such as:
   - abbreviation
   - singular/plural form
   - common short form
9) Do NOT include long explanations, speculative synonyms, or unrelated concepts.
10) If an option contains no meaningful medical entity, return an empty entity list for that option.

Return output in exactly this shape:
{
  "options": [
    {
      "option_key": "A",
      "entities": [
        {
          "text": "...",
          "normalized": "...",
          "variants": ["...", "..."]
        }
      ]
    }
  ]
}

Do NOT return a dictionary keyed by option labels.
Return a list under `options`.

Return structured JSON matching the provided schema exactly.

[paste options]
\end{Verbatim}

\subsection{Differential Evidence Generation}
\begin{Verbatim}[fontsize=\small,breaklines=true,breakanywhere=true,breaksymbolleft={},breaksymbolright={},frame=single,framesep=3pt]
SYSTEM
You are a medical differential-evidence generator for multiple-choice question distillation.

USER
Input includes:
1. the original question,
2. anchor facts extracted from the clinical background,
3. the answer options,
4. co-occurrence paragraphs between anchor facts and options.
Task: For each option, generate one short differential-evidence string that identifies the option's most distinctive clinical decision role in the setting of the question.
Definition of differential evidence: Differential evidence is a short, option-specific clinical cue that places the option on the most relevant decision axis for this question, such as treatment role, timing, mechanism, contraindication, adverse-effect relevance, management category, or typical use setting. It should help distinguish the option conceptually from competing options, but it must not read like an answer explanation or a statement about whether the option fits the case.
Guidelines:
- Use the original question only to infer the general decision setting and the likely clinical management axis.
- Do not restate or reconstruct the full case.
- Use co-occurrence paragraphs as grounding constraints when available.
- If no co-occurrence paragraph is available, use internal medical knowledge conservatively and keep it consistent with the question setting.
- Prefer the option's most distinctive decision role over generic drug summaries, broad mechanism descriptions, or nonessential side effects.
- For medications, prioritize therapeutic role, management timing, treatment category, or clinically important risk over general pharmacology unless pharmacology is itself the key distinguishing cue.
- Keep the evidence concise and focused.
- Write one sentence whenever possible.
Restrictions:
- Do not solve the question.
- Do not say or imply which option is correct, incorrect, preferred, excluded, or should be selected.
- Do not rank options.
- Do not copy the source text verbatim.
- Do not output reasoning steps, bullet points, or extra commentary.
- Do not explain why the option matches or fails to match the current case.
- Do not directly justify the statement using the patient, vignette, described findings, or anchor facts.
- Do not distinguish the option by saying what it is not used for, not part of, or less appropriate than.
- Avoid negative contrast constructions such as: "rather than ...", "instead of ...", "not part of ...", "not used for ...", "less appropriate", "ruled out".
- Express the evidence as a positive description of the option's own clinical decision role.
- Avoid generic textbook descriptions that are not useful for distinguishing the option on the question's decision axis.
Output format:
Return only a JSON object mapping each option label to its differential-evidence string.
Each string should be a standalone concise sentence.
[paste question, anchor cues, options and co-occurrence paragraphs]
\end{Verbatim}

\subsection{Answering with Differential Evidence}
\begin{Verbatim}[fontsize=\small,breaklines=true,breakanywhere=true,breaksymbolleft={},breaksymbolright={},frame=single,framesep=3pt]
SYSTEM
You are taking a medical multiple-choice exam.

USER
Use the question, options, and differential evidence to choose the single best answer.

Differential evidence refers to the most distinguishing evidence for each option, that is, the evidence most useful for separating that option from the competing options.

Use the differential evidence as a set of discriminative clues, and then look back to the question context to identify the facts most consistent with one option over the others.

Do not treat the evidence as a final conclusion by itself; use it to guide context-grounded reasoning.

Reply with only one capital letter: A, B, C, D, E, F, G, H, I, J.

Question: [paste context and question]

Options: [paste options]

Retrieved Evidence: [paste differential evidences]

Answer:
\end{Verbatim}

\subsection{CoT Answering with Retrieved Co-occurrence Text}
\begin{Verbatim}[fontsize=\small,breaklines=true,breakanywhere=true,breaksymbolleft={},breaksymbolright={},frame=single,framesep=3pt]
SYSTEM
You are taking a medical multiple-choice exam.

USER
Use the question, options, and retrieved evidence to reason briefly and choose the single best answer.

The retrieved evidence below consists of raw co-occurrence text associated with the question and candidate options. Use it as supporting context, but do not assume it is conclusive by itself.

Compare the options briefly, ground your reasoning in the question context, and choose the single best answer.
Keep the reasoning short (2-4 sentences). Do not list option letters in the reasoning.

Output format:

Reasoning: <2-4 short sentences>

Final Answer: one capital letter among A, B, C, D, E, F, G, H, I, J.

Question: [paste context and question]

Options: [paste options]

Retrieved Evidence: [paste differential evidences]

Reasoning:
\end{Verbatim}

\end{document}